\documentclass[10pt,twocolumn,letterpaper]{article}
\usepackage[utf8]{inputenc}
\usepackage{cvpr}
\usepackage{times}
\usepackage{graphicx}
\usepackage{amsmath,amssymb} 
\usepackage[breaklinks=true,bookmarks=false]{hyperref}
\usepackage{color}
\usepackage[caption=false, font=footnotesize]{subfig}
\usepackage{cite}

\cvprfinalcopy 

\def\cvprPaperID{0006}

\ifcvprfinal\fi

\begin{document}

\title{An Aggregated Multicolumn Dilated Convolution Network\\ for Perspective-Free Counting}

\author{Diptodip Deb \\
Georgia Institute of Technology\\
{\tt\small diptodipdeb@gatech.edu}
\and
Jonathan Ventura \\
University of Colorado Colorado Springs\\
{\tt\small jventura@uccs.edu}
}

\maketitle

\begin{abstract}
We propose the use of dilated filters to construct an aggregation module in a multicolumn convolutional neural network for perspective-free counting. Counting is a common problem in computer vision (e.g. traffic on the street or pedestrians in a crowd). Modern approaches to the counting problem involve the production of a density map via regression whose integral is equal to the number of objects in the image. However, objects in the image can occur at different scales (e.g. due to perspective effects) which can make it difficult for a learning agent to learn the proper density map. While the use of multiple columns to extract multiscale information from images has been shown before, our approach aggregates the multiscale information gathered by the multicolumn convolutional neural network to improve performance. Our experiments show that our proposed network outperforms the state-of-the-art on many benchmark datasets, and also that using our aggregation module in combination with a higher number of columns is beneficial for multiscale counting.
\end{abstract}


%

\section{Introduction}

Learning to count the number of objects in an image is a deceptively difficult problem with many interesting applications, such as surveillance \cite{ryan2009crowd}, traffic monitoring \cite{TRANCOSdataset_IbPRIA2015} and medical image analysis \cite{selinummi2005software}.  In many of these application areas, the objects to be counted vary widely in appearance, size and shape, and labeled training data is typically sparse. These factors pose a significant computer vision and machine learning challenge.

Lempitsky et al.~\cite{lempitsky2010} showed that it is possible to learn to count without learning to explicitly detect and localize individual objects. Instead, they propose learning to predict a density map whose integral over the image equals the number of objects in the image. This approach has been adopted by many later works (Cf.~\cite{rubio2016,zhang2016single}).


However, in many counting problems, such as those counting cells
in a microscope image, pedestrians in a crowd, or vehicles in a traffic jam, regressors trained on a single image scale are not reliable \cite{rubio2016}. This is due to a variety of challenges including overlap of objects and perspective effects which cause significant variance in object shape, size and appearance.

The most successful recent approaches address this issue by explicitly incorporating multi-scale information in the network \cite{rubio2016,zhang2016single}. These approaches either combine multiple networks which take input patches of different sizes \cite{rubio2016} or combine multiple filtering paths (``columns'') which have different size filters \cite{zhang2016single}.


Following on the intuition that multiscale integration is key to achieving good counting performance, we propose to incorporate dilated filters \cite{yu2015} into a multicolumn convolutional neural network design \cite{zhang2016single}. Dilated filters exponentially increase the network's receptive field without an exponential increase in parameters, allowing for efficient use of multiscale information. Convolutional neural networks with dilated filters have proven to provide competitive performance in image segmentation where multiscale analysis is also critical \cite{yu2015,yu2017dilated}. By incorporating dilated filters into the multicolumn network design, we greatly increase the ability of the network to selectively aggregate multiscale information, without the need for explicit perspective maps during training and testing. We propose the ``aggregated multicolumn dilated convolution network'' or AMDCN which uses dilations to aggregate multiscale information. Our extensive experimental evaluation shows that this proposed network outperforms previous methods on many benchmark datasets.


\begin{figure*}
    \begin{center}
\includegraphics[width=\textwidth]{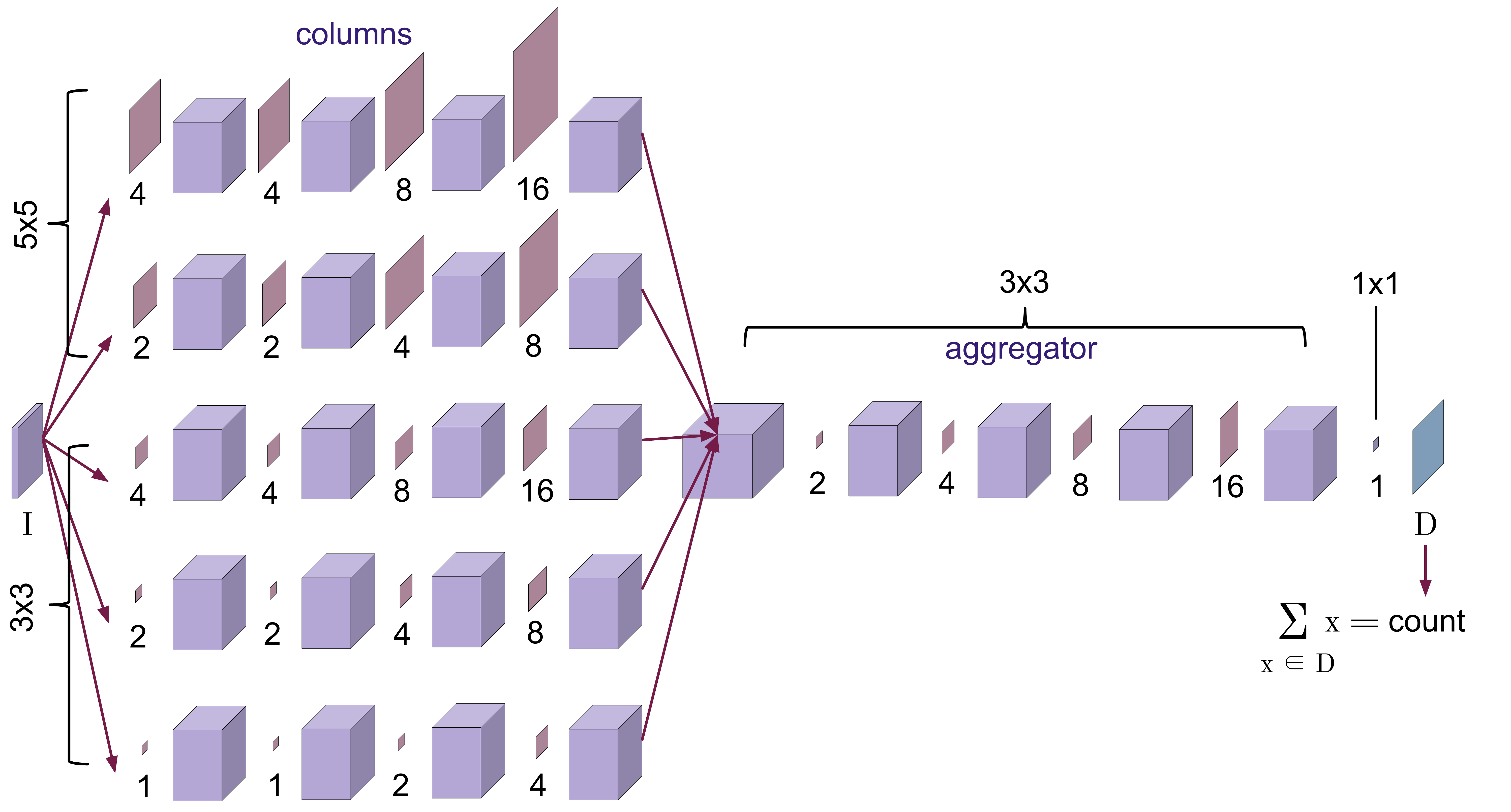}
    \end{center}
    \caption{Fully convolutional architecture diagram (not to scale). Arrows show separate columns that all take the same input. At the end of the columns, the feature maps are merged (concatenated) together and passed to another series of dilated convolutions: the aggregator, which can aggregate the multiscale information collected by the columns \cite{yu2015}. The input image is I with C channels. The output single channel density map is D, and integrating over this map (summing the pixels) results in the final count. Initial filter sizes are labeled with brackets or lines. Convolution operations are shown as flat rectangles, feature maps are shown as prisms. The number below each filter represents the dilation rate (1 means no dilation).}
\label{fig:architecture}
\end{figure*}

\section{Related Work}
Counting using a supervised regressor to formulate a density map was first shown by 
\cite{lempitsky2010}. In this paper, Lempitsky et al. show that the minimal annotation of a 
single dot blurred by a Gaussian kernel produces a sufficient density map to train a network
to count. All of the counting methods that we examine as well as the method we use in our
paper follow this method of producing a density map via regression. This is particularly
advantageous because a sufficiently accurate regressor can also locate the objects in the
image via this method. However, the Lempitsky paper ignores the issue of perspective scaling
and other scaling issues. The work of \cite{zhang2015cross} introduces CNNs (convolutional neural networks) for the purposes of crowd counting, but performs regression on similarly scaled image patches.

These issues are addressed by the work of \cite{rubio2016}. Rubio et al. show that a fully convolutional neural network can be used to produce a supervised regressor that produces density maps as in \cite{lempitsky2010}. They further demonstrate a method dubbed HydraCNN which essentially combines multiple convolutional networks that take in differently scaled image patches in order to incorporate multiscale, global information from the image. The premise of this method is that a single regressor will fail to accurately represent the difference in values of the features of an image caused by perspective shifts (scaling effects) \cite{rubio2016}.

However, the architectures of both \cite{rubio2016} and \cite{zhang2015cross} are not fully convolutional due to requiring multiple image patches and, as discussed in \cite{yu2015}, the experiments of \cite{noh2015, fischer2015} and \cite{farabet2013, lin2016, chen2016} leave it unclear as to whether rescaling patches of the image is truly necessary in order to solve dense prediction problems via convolutional neural networks. Moreover, these approaches seem to saturate in performance at three columns, which means the network is extracting information from fewer scales. The work of \cite{yu2015} proposes the use of dilated convolutions as a simpler alternative that does not require sampling of rescaled image patches to provide global, scale-aware information to the network. A fully convolutional approach to multiscale counting has been proposed by \cite{zhang2016single}, in which a multicolumn convolutional network gathers features of different scales by using convolutions of increasing kernel sizes from column to column instead of scaling image patches. Further, DeepLab has used dilated convolutions in multiple columns to extract scale information for segmentation \cite{chen2017deeplab}. We build on these approaches with our aggregator module as described in Section \ref{sec:method}, which should allow for extracting information from more scales.

It should be noted that other methods of counting exist, including training a network to recognize deep object features via only providing the counts of the objects of interest in an image \cite{segui2015} and using CNNs (convolutional neural networks) along with boosting in order to improve the results of regression for production of density maps \cite{walach2016}. In the same spirit, \cite{boominathan2016} combines deep and shallow convolutions within the same network, providing accurate counting of dense objects (e.g. the UCF50 crowd dataset).

In this paper, however, we aim to apply the dilated convolution method of \cite{yu2015}, which has shown to be able to incorporate multiscale perspective information without using multiple inputs or a complicated network architecture, as well as the multicolumn approach of \cite{zhang2016single, chen2017deeplab} to aggregate multiscale information for the counting problem.

\section{Method}
\subsection{Dilated Convolutions for Multicolumn Networks}\label{sec:method}


\begin{figure}
    \begin{center}
\includegraphics[width=\columnwidth]{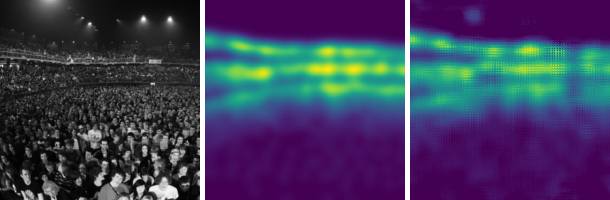}%
    \end{center}
\caption{UCF sample results. Left: input counting image. Middle: Ground truth density map. Right: AMDCN prediction of density map on test image. The network never saw these images during training. All density maps are one channel only (i.e. grayscale), but are colored here for clarity.}
\label{fig:sample_output}
\end{figure}

We propose the use of dilated convolutions as an attractive alternative to the architecture of the HydraCNN \cite{rubio2016}, which seems to saturate in performance at 3 or more columns. We refer to our proposed network as the aggregated multicolumn dilated convolution network\footnote{Implementation available on \url{https://github.com/diptodip/counting}.}, henceforth shortened as the AMDCN. The architecture of the AMDCN is inspired by the multicolumn counting network of \cite{zhang2016single}. Extracting features from multiple scales is a good idea when attempting to perform perspective-free counting and increasing the convolution kernel size across columns is an efficient method of doing so. However, the number of parameters increases exponentially as larger kernels are used in these columns to extract features at larger scales. Therefore, we propose using dilated convolutions rather than larger kernels.

Dilated convolutions, as discussed in \cite{yu2015}, allow for the exponential increase of the receptive field with a linear increase in the number of parameters with respect to each hidden layer.

In a traditional 2D convolution, we define a real valued function \(F: \mathbb{Z}^2 \to \mathbb{R}\), an input \(\Omega_r = [-r, r]^2 \in \mathbb{Z}^2\), and a filter function \(k: \Omega_r \to \mathbb{R}\). In this case, a convolution operation as defined in \cite{yu2015} is given by
\begin{equation}
(F * k)(\mathbf{p}) = \sum_{\mathbf{s} + \mathbf{t} = \mathbf{p}} F(\mathbf{s})k(\mathbf{t}).
\end{equation}

A dilated convolution is essentially a generalization of the traditional 2D convolution that allows the operation to skip some inputs. This enables an increase in the size of the filter (i.e. the size of the receptive field) without losing resolution. Formally, we define from \cite{yu2015} the dilated convolution as 
\begin{equation}(F *_l k)(\mathbf{p}) = \sum_{\mathbf{s} + l\mathbf{t} = \mathbf{p}} F(\mathbf{s})k(\mathbf{t})
\end{equation}
where \(l\) is the index of the current layer of the convolution.

Using dilations to construct the aggregator in combination with the multicolumn idea will allow for the construction of a network with more than just 3 or 4 columns as in \cite{zhang2016single} and \cite{chen2017deeplab}, because the aggregator should prevent the saturation of performance with increasing numbers of columns. Therefore the network will be able to extract useful features from more scales. We take advantage of dilations within the columns as well to provide large receptive fields with fewer parameters.

Looking at more scales should allow for more accurate regression of the density map. However, because not all scales will be relevant, we extend the network beyond a simple \(1 \times 1\) convolution after the merged columns. Instead, we construct a second part of the network, the aggregator, which sets our method apart from \cite{zhang2016single}, \cite{chen2017deeplab}, and other multicolumn networks. This aggregator is another series of dilated convolutions that should appropriately consolidate the multiscale information collected by the columns. This is a capability of dilated convolutions observed by \cite{yu2015}. While papers such as \cite{zhang2016single} and \cite{chen2017deeplab} have shown that multiple columns and dilated columns are useful in extracting multiscale information, we argue in this paper that the simple aggregator module built using dilated convolutions is able to effectively make use multiscale information from multiple columns. We show compelling evidence for these claims in Section \ref{sec:ablation_studies}.

The network as shown in Figure \ref{fig:architecture} contains 5 columns. Note that dilations allow us to use more columns for counting than \cite{zhang2016single} or \cite{chen2017deeplab}. Each column looks at a larger scale than the previous (the exact dilations can also be seen in Figure \ref{fig:architecture}). There are 32 feature maps for each convolution, and all inputs are zero padded prior to each convolution in order to maintain the same data shape from input to output. That is, an image input to this network will result in a density map of the same dimensions. All activations in the specified network are ReLUs. Our input pixel values are floating point 32 bit values from 0 to 1. We center our inputs at 0 by subtracting the per channel mean from each channel. When training, we use a scaled mean absolute error for our loss function:
\begin{equation}
L = \frac{1}{n}\sum_{i=1}^n\lvert \hat{y}_i - \gamma y_i \rvert
\end{equation}
where \(\gamma\) is the scale factor, \(\hat{y}_i\) is the prediction, \(y_i\) is the true value, and \(n\) is the number of pixels. We use a scaled mean absolute error because the target values are so small that it is numerically unstable to regress to these values. At testing time, when retrieving the output density map from the network, we scale the pixel values by \(\gamma^{-1}\) to obtain the correct value. This approach is more numerically stable and avoids having the network learn to output only zeros by weighting the nonzero values highly. For all our datasets, we set \(\gamma = 255\).

\subsection{Experiments}

We evaluated the performance of dilated convolutions against various counting methods on a variety of common counting datasets: UCF50 crowd data, TRANCOS traffic data \cite{rubio2016}, UCSD crowd data \cite{chan2008privacy}, and WorldExpo crowd data \cite{zhang2015cross}. For each of these data sets, we used labels given by the corresponding density map for each image. An example of this is shown in Figure \ref{fig:sample_output}. We have performed experiments on the four different splits of the UCSD data as used in \cite{rubio2016} and the split of the UCSD data as used in \cite{zhang2016single} (which we call the original split). We also evaluated the performance of our network on the TRANCOS traffic dataset \cite{TRANCOSdataset_IbPRIA2015}. We have also experimented with higher density datasets for crowd counting, namely WorldExpo and UCF.

We have observed that multicolumn dilations produce density maps (and therefore counts) that often have lower loss than those of HydraCNN \cite{rubio2016} and \cite{zhang2016single}. We measure density map regression loss via a scaled mean absolute error loss during training. We compare accuracy of the counts via mean absolute error for the crowd datasets and the GAME metric in the TRANCOS dataset as explained in Section \ref{trancos_explanation}. Beyond the comparison to HydraCNN, we will also compare to other recent convolutional counting methods, especially those of \cite{segui2015}, \cite{walach2016}, and \cite{boominathan2016} where possible.

For all datasets, we generally use patched input images and ground truth density maps produced by summing a Gaussian of a fixed size (\(\sigma\)) for each object for training. This size varies from dataset to dataset, but remains constant within a dataset with the exception of cases in which a perspective map is used. This is explained per dataset. All experiments were performed using Keras with the Adam optimizer \cite{chollet2015keras}. The learning rates used are detailed per dataset. For testing, we also use patches that can either be directly pieced together or overlapped and averaged except in the case of UCF, for which we run our network on the full image.

Furthermore, we performed a set of experiments in which we varied the number of columns from 1 to 5 (simply by including or not including the columns as specified in Figure \ref{fig:architecture}, starting with the smallest filter column and adding larger filter columns one by one). Essentially, the network is allowed to extract information at larger and larger scales in addition to the smaller scales as we include each column. We then performed the same set of experiments, varying the number of columns, but with the aggregator module removed. We perform these experiments on the original split of UCSD as specified in Section \ref{sec:ucsd_explanation} and \cite{chan2008privacy}, the TRANCOS dataset, and the WorldExpo dataset because these are relatively large and well defined datasets. We limit the number of epochs to 10 for all of these sets of experiments in order to control for the effect of learning time, and also compare all results using MAE for consistency. These experiments are key to determining the efficacy of the aggregator in effectively combining multiscale information and in providing evidence to support the use of multiple columns to extract multiscale information from images. We report the results of these ablation studies in Section \ref{sec:ablation_studies}.

\subsubsection{UCF50 Crowd Counting}

UCF is a particularly challenging crowd counting dataset. There are only 50 images in the whole dataset and they are all of varying sizes and from different scenes. The number of people also varies between images from less than 100 to the thousands. The average image has on the order of 1000 people. The difficulty is due to the combination of the very low number of images in the dataset and the fact that the images are all of varying scenes, making high quality generalization crucial. Furthermore, perspective effects are particularly noticeable for many images in this dataset. Despite this, there is no perspective information available for this dataset.

We take 1600 random patches of size \(150 \times 150\) for the training. For testing, we do not densely scan the image as in \cite{rubio2016} but instead test on the whole image. In order to standardize the image sizes, we pad each image out with zeros until all images are \(1024 \times 1024\). We then suppress output in the regions where we added padding when testing. This provides a cleaner resulting density map for these large crowds. The ground truth density maps are produced by annotating each object with a Gaussian of \(\sigma = 15\).

\subsubsection{TRANCOS Traffic Counting}\label{trancos_explanation}

TRANCOS is a traffic counting dataset that comes with its own metric \cite{TRANCOSdataset_IbPRIA2015}. This metric is known as \(GAME\), which stands for Grid Average Mean absolute Error. \(GAME\) splits a given density map into \(4^L\) grids, or subarrays, and obtains a mean absolute error within each grid separately. The value of \(L\) is a parameter chosen by the user. These individual errors are summed to obtain the final error for a particular image. The intuition behind this metric is that it is desirable to penalize a density map whose overall count might match the ground truth, but whose shape does not match the ground truth \cite{TRANCOSdataset_IbPRIA2015}. More formally, we define
\begin{equation}
GAME(L) = \frac{1}{N} \cdot \sum_{n = 1}^{N} \left( \sum_{l = 1}^{4^L}\lvert e_n^l - t_n^l \rvert \right)
\end{equation}
where \(N\) refers to the number of images, \(L\) is the level parameter for \(GAME\), \(e_n^l\) is the predicted or estimated count in region \(l\) of image \(n\) and \(t_n^l\) is the ground truth count in region \(l\) of image \(n\) \cite{TRANCOSdataset_IbPRIA2015}.

For training this dataset, we take 1600 randomly sampled patches of size \(80 \times 80\). For testing this dataset, we take \(80 \times 80\) non-overlapping patches which we can stitch back together into the full-sized \(640 \times 480\) images. We trained the AMDCN network with density maps produced with a Gaussian of \(\sigma = 15\) as specified in \cite{rubio2016}.

\subsubsection{UCSD Crowd Counting}\label{sec:ucsd_explanation}

The UCSD crowd counting dataset consists of frames of video of a sidewalk. There are relatively few people in view at any given time (approximately 25 on average). Furthermore, because the dataset comes from a video, there are many nearly identical images in the dataset. For this dataset, there have been two different ways to split the data into train and test sets. Therefore, we report results using both methods of splitting the data. The first method consists of four different splits: maximal, downscale, upscale, and minimal. Minimal is particularly challenging as the train set contains only 10 images. Moreover, upscale appears to be the easiest for the majority of methods \cite{rubio2016}. The second method of splitting this data is much more succinct, leaving 1200 images in the testing set and 800 images in the training set \cite{zhang2016single}. This split comes from the original paper, so we call it the original split \cite{chan2008privacy}.

For this dataset, each object is annotated with a 2D Gaussian of covariance \(\Sigma = 8\cdot \mathbf{1}_{2 \times 2}\). The ground truth map is produced by summing these. When we make use of the perspective maps provided, we divide \(\Sigma\) by the perspective map value at that pixel \(\mathbf{x}\), represented by \(M(\mathbf{x})\). The provided perspective map for UCSD contains both a horizontal and vertical direction so we take the square root of the provided combined value. For training, we take 1600 random \(79 \times 119\) pixel patches and for testing, we split each test image up into quadrants (which have dimension \(79 \times 119\)). There are two different ways to split the dataset into training and testing sets. We have experimented on the split that gave \cite{rubio2016} the best results as well as the split used in \cite{zhang2016single}.

First, we split the dataset into four separate groups of training and testing sets as used in \cite{rubio2016} and originally defined by \cite{ryan2009crowd}. These groups are ``upscale,'' ``maximal,'' ``minimal,'' and ``downscale.'' We see in Table \ref{tab:ucsd} that the ``upscale'' split and ``downscale'' split give us state of the art results on counting for this dataset. For this experiment, we sampled 1600 random patches of size \(119 \times 79\) pixels (width and height respectively) for the training set and split the test set images into \(119 \times 79\) quadrants that could be reconstructed by piecing them together without overlap. We also added left-right flips of each image to our training data.

We then evaluate the original split. For this experiment, we similarly sampled 1600 random patches of size \(119 \times 79\) pixels (width and height respectively) for the training set and split the test set images into \(119 \times 79\) quadrants that could be reconstructed by piecing them together without overlap.

\subsubsection{WorldExpo '10 Crowd Counting}

The WorldExpo dataset \cite{zhang2015cross} contains a larger number of people (approximately 50 on average, which is double that of UCSD) and contains images from multiple locations. Perspective effects are also much more noticeable in this dataset as compared to UCSD. These qualities of the dataset serve to increase the difficulty of counting. Like UCSD, the WorldExpo dataset was constructed from frames of video recordings of crowds. This means that, unlike UCF, this dataset contains a relatively large number of training and testing images. We experiment on this dataset with and without perspective information.

Without perspective maps, we generate label density maps for this dataset in the same manner as previously described: a 2D Gaussian with \(\sigma = 15\). We take 16000 \(150 \times 150\) randomly sampled patches for training. For testing, we densely scan the image, producing \(150 \times 150\) patches at a stride of 100.

When perspective maps are used, however, we follow the procedure as described in \cite{zhang2015cross}, which involves estimating a ``crowd density distribution kernel'' as the sum of two 2D Gaussians: a symmetric Gaussian for the head and an ellipsoid Gaussian for the body. These are scaled by the perspective map \(M\) provided, where \(M(\mathbf{x})\) gives the number of pixels that represents a meter at pixel \(\mathbf{x}\) \cite{zhang2015cross}. Note that the meaning of this perspective map is distinct from the meaning of the perspective map provided for the UCSD dataset. Using this information, the density contribution from a person with head pixel \(\mathbf{x}\) is given by the following sum of normalized Gaussians:
\begin{equation}
    D_\mathbf{x} = \frac{1}{\lvert \lvert Z \rvert \rvert }(\mathcal{N}_h(\mathbf{x}, \sigma_h) + \mathcal{N}_b(\mathbf{x}_b, \Sigma_b)) 
\end{equation}
where \(\mathbf{x}_b\) is the center of the body, which is 0.875 meters down from the head on average, and can be determined from the perspective map \(M\) and the head center \(\mathbf{x}\) \cite{zhang2015cross}.
We sum these Gaussians for each person to produce the final density map. We set \(\sigma = 0.2M(\mathbf{x})\) for \(\mathcal{N}_h\) and \(\sigma_x = 0.2M(\mathbf{x}), \sigma_y = 0.5M(\mathbf{x})\) for \(\Sigma_b\) in \(\mathcal{N}_b\).

\section{Results}

\subsection{UCF Crowd Counting}\label{ucf_50}

The UCF dataset is particularly challenging due to the large number of people in the images, the variety of the scenes, as well as the low number of training images. We see in Figure \ref{fig:sample_output} that because the UCF dataset has over 1000 people on average in each image, the shapes output by the network in the density map are not as well defined or separated as in the UCSD dataset.

We report a state of the art result on this dataset in Table \ref{tab:ucf}, following the standard protocol of 5-fold cross validation. Our MAE on the dataset is 290.82, which is approximately 5 lower than the previous state of the art, HydraCNN \cite{rubio2016}. This is particularly indicative of the power of an aggregated multicolumn dilation network. Despite not making use of perspective information, the AMDCN is still able to produce highly accurate density maps for UCF.

\begin{table}
    \begin{center}
    \begin{tabular}{|p{3.5cm}|p{2.5cm}|}
        \hline
         \textbf{Method} & \textbf{MAE}\\\hline
         AMDCN & \textbf{290.82}\\\hline
         Hydra2s \cite{rubio2016} & 333.73\\\hline
         MCNN \cite{zhang2016single} & 377.60\\\hline
         \cite{zhang2015cross} & 467.00\\\hline
         \cite{sindagi2017generating} & 295.80\\\hline
         \cite{babu2017switching} & 318.10\\\hline
    \end{tabular}
    \end{center}
    \caption{Mean absolute error of various methods on UCF crowds}
    \label{tab:ucf}
\end{table}

\subsection{TRANCOS Traffic Counting}\label{trancos}

Our network performs very well on the TRANCOS dataset. Indeed, as confirmed by the GAME score, AMDCN produces the most accurate count and shape combined as compared to other methods. Table \ref{tab:trancos} shows that we achieve state of the art results as measured by the \(GAME\) metric \cite{TRANCOSdataset_IbPRIA2015} across all levels.

\begin{table}
    \begin{center}
    \begin{tabular}{|p{2.25cm}|p{1cm}|p{1cm}|p{1cm}|p{1cm}|}
        \hline
         \textbf{Method} & \textbf{GAME (L=0)} & \textbf{GAME (L=1)} & \textbf{GAME (L=2)} & \textbf{GAME (L=3)}\\\hline
         AMDCN & \textbf{9.77} & \textbf{13.16} & \textbf{15.00} & \textbf{15.87}\\\hline
         \cite{rubio2016} & 10.99 & 13.75 & 16.69 & 19.32\\\hline
         \cite{lempitsky2010} + SIFT from \cite{TRANCOSdataset_IbPRIA2015} & 13.76 & 16.72 & 20.72 & 24.36\\\hline
         \cite{fiaschi2012learning} + RGB Norm + Filters from \cite{TRANCOSdataset_IbPRIA2015} & 17.68 & 19.97 & 23.54 & 25.84\\\hline
         HOG-2 from \cite{TRANCOSdataset_IbPRIA2015} & 13.29 & 18.05 & 23.65 & 28.41\\\hline
    \end{tabular}
    \end{center}
    \caption{Mean absolute error of various methods on TRANCOS traffic}
    \label{tab:trancos}
\end{table}

\subsection{UCSD Crowd Counting}

\begin{figure*}
    \begin{center}
        \subfloat[UCSD upscale split.]{\includegraphics[width=0.45\textwidth]{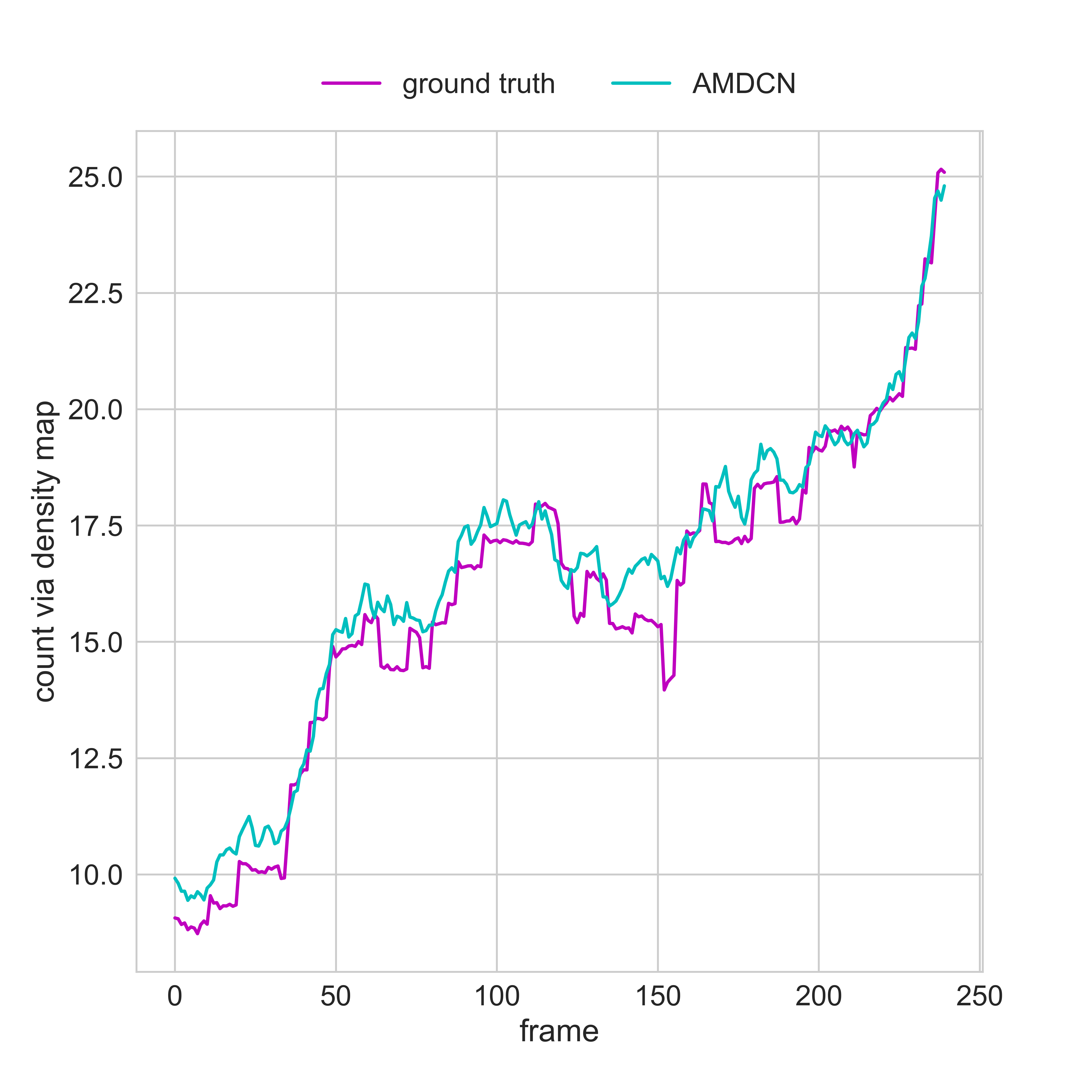}}%
\hfil
        \subfloat[UCSD original split.]{\includegraphics[width=0.45\textwidth]{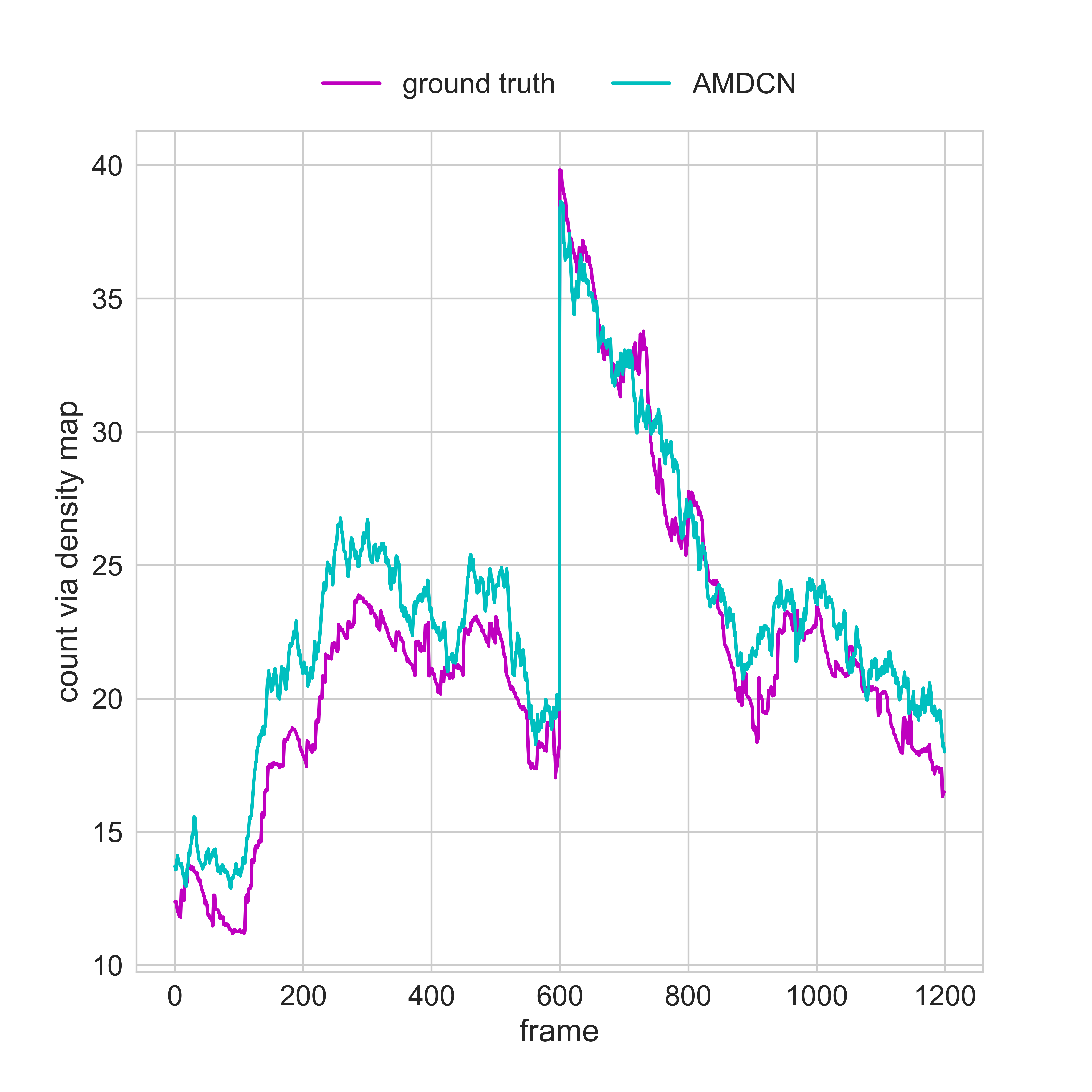}}%
\hfil
    \end{center}
\caption{UCSD crowd counting dataset. Both plots show comparisons of predicted and ground truth counts over time. While AMDCN does not beat the state of the art on the original split, the predictions still follow the true counts reasonably. The jump in the original split is due to that testing set including multiple scenes of highly varying counts.}
\label{fig:ucsd_plots}
\end{figure*}

Results are shown in Table \ref{tab:ucsd} and Figure \ref{fig:ucsd_plots}. We see that the ``original'' split as defined by the creators of the dataset in \cite{chan2008privacy} and used in \cite{zhang2016single} gives us somewhat worse results for counting on this dataset. Results were consistent over multiple trainings. Again, including the perspective map does not seem to increase performance on this dataset. Despite this, we see in Table \ref{tab:ucsd} and Figure \ref{fig:ucsd_plots} that the results are comparable to the state of the art. In fact, for two of the splits, our proposed network beats the state of the art. For the upscale split, the AMDCN is the state of the art by a large relative margin. This is compelling because it shows that accurate perspective-free counting can be achieved without creating image pyramids or requiring perspective maps as labels using the techniques presented by the AMDCN.

\begin{table*}
    \begin{center}
    \begin{tabular}{|p{6.5cm}|p{1.5cm}|p{1.6cm}|p{1.5cm}|p{1.5cm}|p{1.5cm}|}
        \hline
         \textbf{Method} & \textbf{maximal} & \textbf{downscale} & \textbf{upscale} & \textbf{minimal} & \textbf{original}\\\hline
         AMDCN \textbf{(without perspective information)} & 1.63 & 1.43 & \textbf{0.63} & 1.71 & 1.74\\\hline
         AMDCN (with perspective information) & 1.60 & \textbf{1.24} & 1.37 & 1.59 & 1.72\\\hline
         \cite{rubio2016} (with perspective information) & 1.65 & 1.79 & 1.11 & 1.50 & -\\\hline
         \cite{rubio2016} (without perspective information) & 2.22 & 1.93 & 1.37 & 2.38 & -\\\hline
         \cite{lempitsky2010} & 1.70 & 1.28 & 1.59 & 2.02 & -\\\hline
         \cite{fiaschi2012learning} & 1.70 & 2.16 & 1.61 & 2.20 & -\\\hline
         \cite{pham2015count} & 1.43 & 1.30 & 1.59 & 1.62 & -\\\hline
         \cite{arteta2014interactive} & \textbf{1.24} & 1.31 & 1.69 & \textbf{1.49} & -\\\hline
         \cite{zhang2015cross} & 1.70 & 1.26 & 1.59 & 1.52 & 1.60\\\hline
         \cite{zhang2016single} & - & - & - & - & \textbf{1.07}\\\hline
         \cite{an2007face, zhang2016single} & - & - & - & - & 2.16\\\hline
         \cite{chen2012feature} & - & - & - & - & 2.25\\\hline
         \cite{chan2008privacy} & - & - & - & - & 2.24\\\hline
         \cite{chen2013cumulative} & - & - & - & - & 2.07\\\hline
    \end{tabular}
    \end{center}
    \caption{Mean absolute error of various methods on UCSD crowds}
    \label{tab:ucsd}
\end{table*}

\subsection{WorldExpo '10 Crowd Counting}\label{worldexpo}


Our network performs reasonably well on the more challenging WorldExpo dataset. While it does not beat the state of the art, our results are comparable. What is more, we do not need to use the perspective maps to obtain these results. As seen in Table \ref{tab:worldexpo}, the AMDCN is capable of incorporating the perspective effects without scaling the Gaussians with perspective information. This shows that it is possible to achieve counting results that approach the state of the art with much simpler labels for the counting training data.

\begin{table}
    \begin{center}
    \begin{tabular}{|p{5.5cm}|p{0.7cm}|}
        \hline
         \textbf{Method} & \textbf{MAE}\\\hline
         AMDCN \textbf{(without perspective information)} & 16.6\\\hline
         AMDCN (with perspective information) & 14.9\\\hline
         LBP+RR \cite{zhang2016single} (with perspective information) & 31.0\\\hline
         MCNN \cite{zhang2016single} (with perspective information) & \textbf{11.6}\\\hline
         \cite{zhang2015cross} (with perspective information) & 12.9\\\hline
    \end{tabular}
    \end{center}
    \caption{Mean absolute error of various methods on WorldExpo crowds}
    \label{tab:worldexpo}
\end{table}

\subsection{Ablation Studies}\label{sec:ablation_studies}

\begin{figure*}[!t]
    \begin{center}
        \subfloat[WorldExpo]{\includegraphics[width=0.32\textwidth]{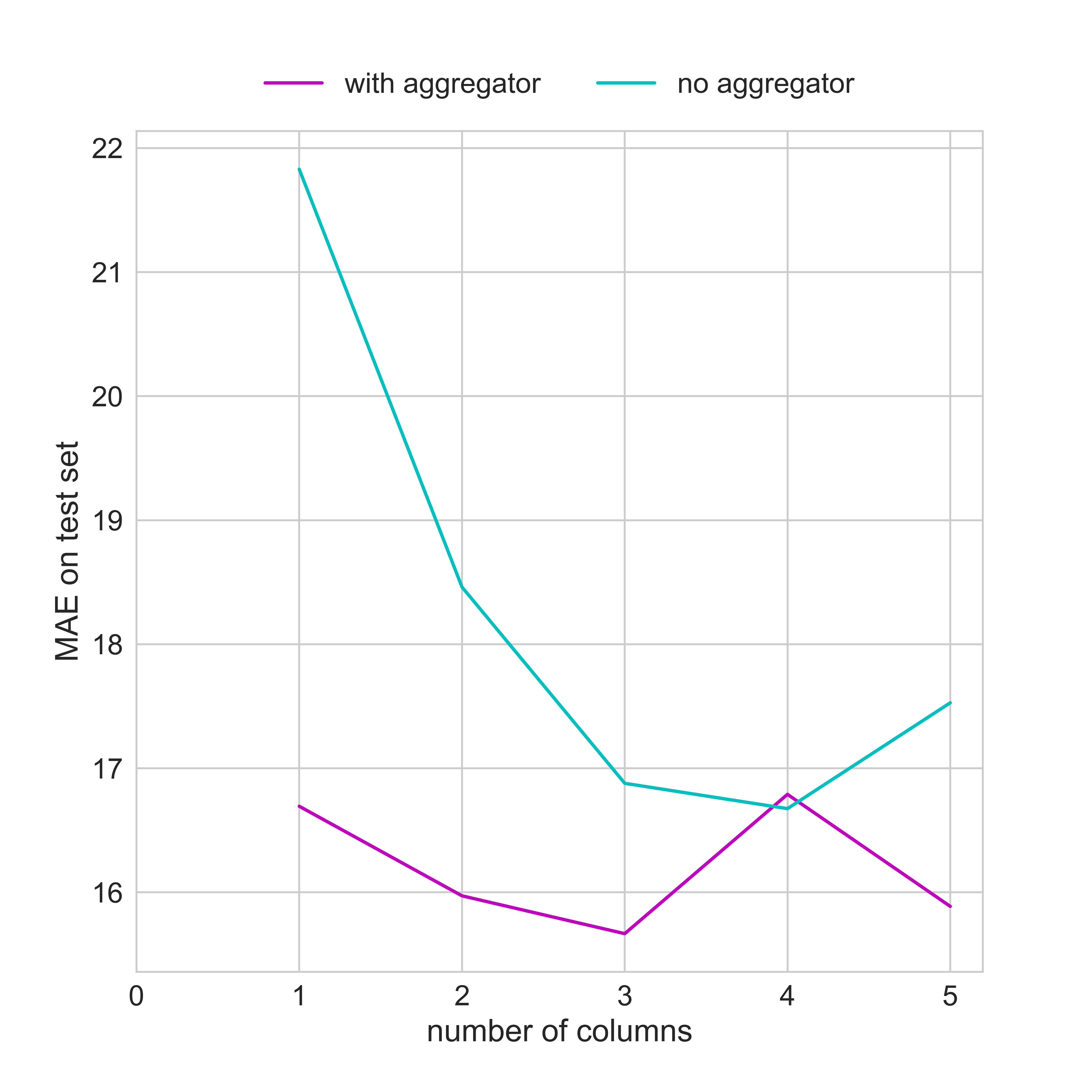}}%
\hfil
        \subfloat[TRANCOS]{\includegraphics[width=0.32\textwidth]{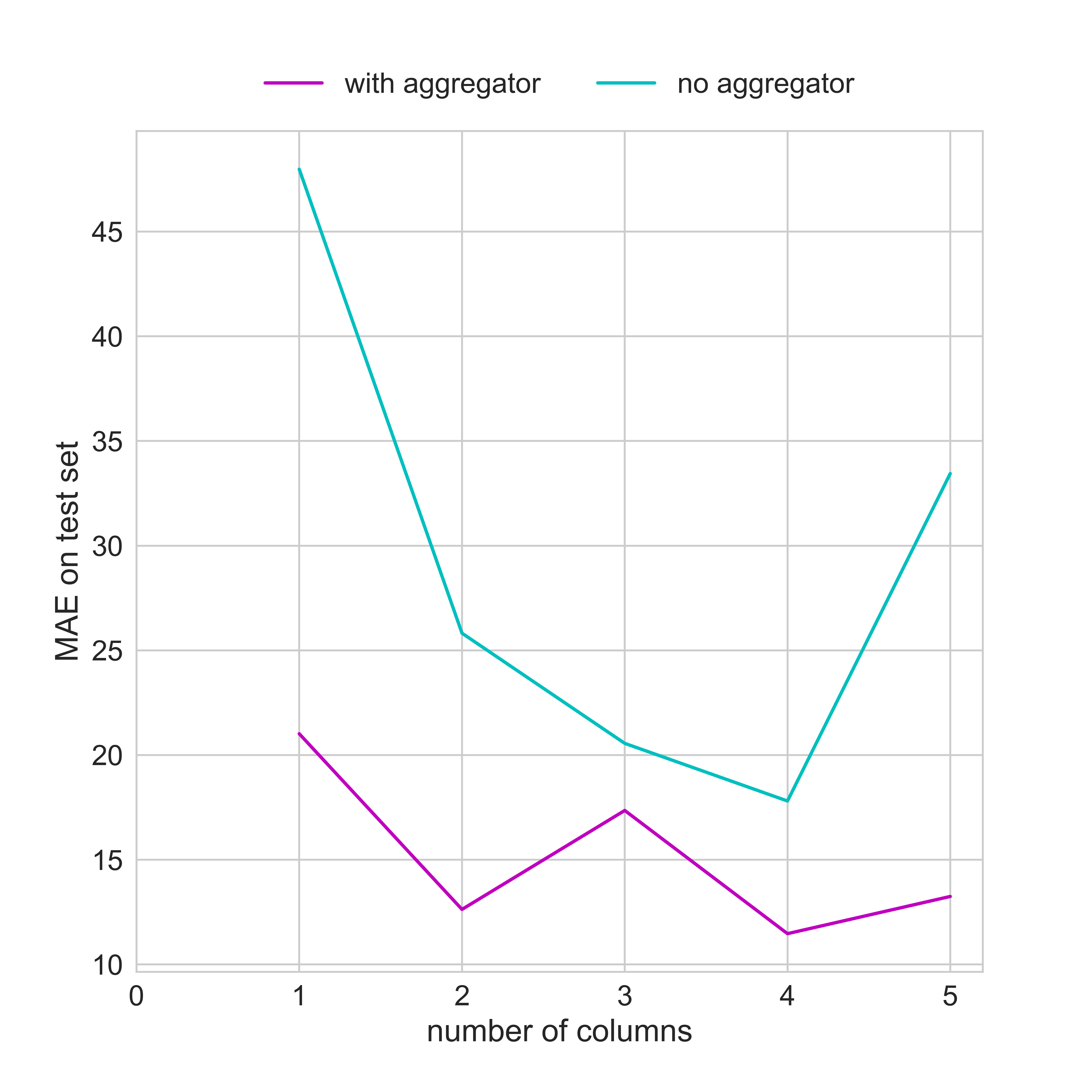}}%
\hfil
        \subfloat[UCSD original split]{\includegraphics[width=0.32\textwidth]{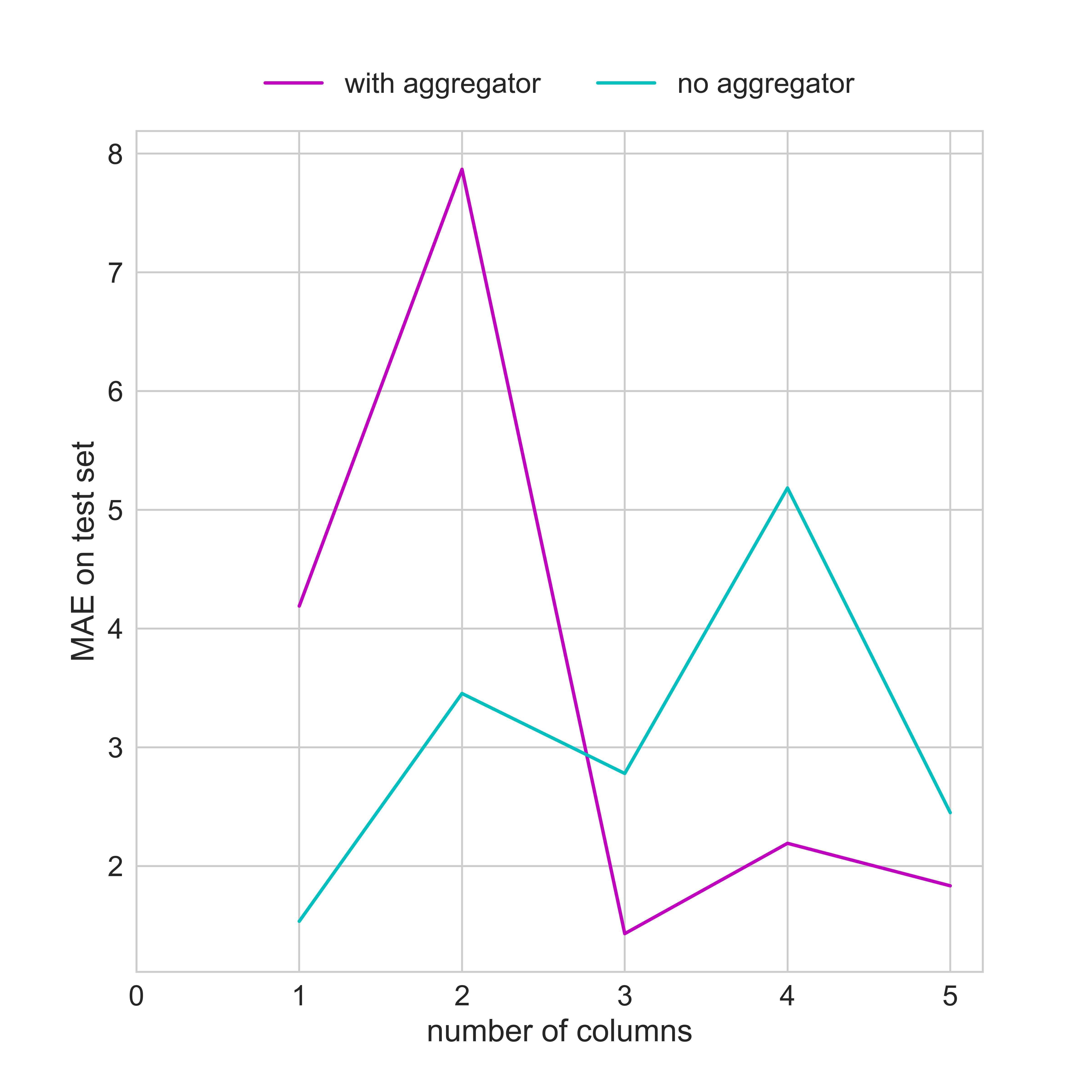}}%
\end{center}
\caption{Ablation studies on various datasets in which the number of columns is varied and the aggregator is included or not included. The results generally support the use of more columns and an aggregator module.}
\label{fig:ablation_studies}
\end{figure*}

We report the results of the ablation studies in Figure \ref{fig:ablation_studies}. We note from these plots that while there is variation in performance, a few trends stand out. Most importantly, the lowest errors are consistently with a combination of a larger number of columns and including the aggregator module. Notably for the TRANCOS dataset, including the aggregator consistently improves performance. Generally, the aggregator tends to decrease the variance in performance of the network. Some of the variance that we see in the plots can be explained by: (1) for lower numbers of columns, including an aggregator is not as likely to help as there is not much separation of multiscale information across columns and (2) for the UCSD dataset, there is less of a perspective effect than TRANCOS and WorldExpo so a simpler network is more likely to perform comparably to a larger network. These results verify the notion that using more columns increases accuracy, and also support our justification for the use of the aggregator module.

\section{Conclusion}
\subsection{Summary}
We have proposed the use of aggregated multicolumn dilated convolutions, the AMDCN, as an alternative to the HydraCNN \cite{rubio2016} or multicolumn CNN \cite{zhang2016single} for the vision task of counting objects in images. Inspired by the multicolumn approach to multiscale problems, we also employ dilations to increase the receptive field of our columns. We then aggregate this multiscale information using another series of dilated convolutions to enable a wide network and detect features at more scales. This method takes advantage of the ability of dilated convolutions to provide exponentially increasing receptive fields. We have performed experiments on the challenging UCF crowd counting dataset, the TRANCOS traffic dataset, multiple splits of the UCSD crowd counting dataset, and the WorldExpo crowd counting dataset.

We obtain superior or comparable results in most of these datasets. The AMDCN is capable of outperforming these approaches completely especially when perspective information is not provided, as in UCF and TRANCOS. These results show that the AMDCN performs surprisingly well and is also robust to scale effects. Further, our ablation study of removing the aggregator network shows that using more columns and an aggregator provides the best accuracy for counting --- especially so when there is no perspective information.

\subsection{Future Work}

In addition to an analysis of performance on counting, a density regressor can also be used to locate objects in the image. As mentioned previously, if the regressor is accurate and precise enough, the resulting density map can be used to locate the objects in the image. We expect that in order to do this, one must regress each object to a single point rather than a region specified by a Gaussian. Perhaps this might be accomplished by applying non-maxima suppression to the final layer activations.

Indeed, the method of applying dilated filters to a multicolumn convolutional network in order to enable extracting features of a large number of scales can be applied to various other dense prediction tasks, such as object segmentation at multiple scales or single image depth map prediction. Though we have only conducted experiments on counting and used 5 columns, the architecture presented can be extended and adapted to a variety of tasks that require information at multiple scales.


%


\section*{Acknowledgment}

This material is based upon work supported by the National
Science  Foundation  under  Grant  No. 1359275  and  1659788. Any  opinions,  findings,  and  conclusions  or  recommendations expressed in this material are those of the authors
and do not necessarily reflect the views of the National Science
Foundation.
Furthermore,  we  acknowledge Kyle Yee and Sridhama Prakhya for their helpful conversations and insights during the
research process.




\bibliographystyle{ieee}
%
\bibliography{bibliography}

%





\end{document}